\documentclass{article}
\usepackage{spconf,amsmath,graphicx}

\usepackage{amsmath,amssymb,amsfonts}
\usepackage{array}
\usepackage[caption=false,font=normalsize,labelfont=sf,textfont=sf]{subfig}
\usepackage{textcomp}
\usepackage{stfloats}
\usepackage{url}
\usepackage{verbatim}
\usepackage{bm}
\usepackage{multirow}
\usepackage{booktabs}
\usepackage{siunitx}
\usepackage{algorithm}
\usepackage{algpseudocode}
\usepackage{graphicx}
\usepackage{cite}
\usepackage{setspace}
\usepackage{hyperref}
\hypersetup{hidelinks}
\usepackage{makecell}
\usepackage{graphicx}

\usepackage[capitalize]{cleveref}
\crefname{section}{Sec.}{Secs.}
\Crefname{section}{Section}{Sections}
\Crefname{table}{Table}{Tables}
\crefname{table}{Tab.}{Tabs.}

\title{PIVM: Diffusion-Based Prior-Integrated Variation Modeling for Anatomically Precise Abdominal CT Synthesis}
\name{
Dinglun He$^{1}$\sthanks{These authors contributed equally to this work.}\hspace{0.35cm}
Baoming Zhang$^{2}$\footnotemark[1]\hspace{0.35cm}
Xu Wang$^{1}$\footnotemark[1]\hspace{0.35cm}
Yao Hao$^{3}$\hspace{0.35cm}
Deshan Yang$^{4}$\hspace{0.35cm}
Ye Duan$^{5}$
}

\address{
$^{1}$ Department of Electrical Engineering and Computer Science, University of Missouri, Columbia, MO, USA\\
$^{2}$ Department of Computer Science, The University of Texas at Dallas, Richardson, TX, USA\\
$^{3}$ Department of Radiation Oncology, Washington University in St. Louis, MO, USA\\
$^{4}$ Department of Radiation Oncology, Duke University, Durham, NC, USA\\
$^{5}$ School of Computing, Clemson University, Clemson, SC, USA
}

\begin{document}
%
\maketitle

\begin{abstract}
Abdominal CT data are limited by high annotation costs and privacy constraints, which hinder the development of robust segmentation and diagnostic models. We present a Prior-Integrated Variation Modeling (PIVM) framework, a diffusion-based method for anatomically accurate CT image synthesis. Instead of generating full images from noise, PIVM predicts voxel-wise intensity variations relative to organ-specific intensity priors derived from segmentation labels. These priors and labels jointly guide the diffusion process, ensuring spatial alignment and realistic organ boundaries. Unlike latent-space diffusion models, our approach operates directly in image space while preserving the full Hounsfield Unit (HU) range, capturing fine anatomical textures without smoothing. Source code is available at \href{https://github.com/BZNR3/PIVM}{this URL}.
\end{abstract}


\section{Introduction}
\label{sec:intro}

Abdominal computed tomography (CT) provides rich anatomical information essential for diagnosis and treatment planning~\cite{litjens2017survey,de2022ct,rhee2021role}. However, developing robust deep learning models for abdominal CT remains challenging due to the scarcity of annotated datasets and the constraints of patient privacy. Manual delineation of organs is time-consuming and often inconsistent across institutions~\cite{ma2021abdomenct,deheyab2022overview}, leading to limited data diversity and generalization. These factors have motivated a growing interest in generating anatomically accurate synthetic CT images to supplement real data and improve model robustness.

Earlier approaches relied on Generative Adversarial Networks (GANs)~\cite{goodfellow2014generative,mao2017least,gulrajani2017improved,zhu2017unpaired}, which enabled realistic image synthesis for data augmentation and modality translation~\cite{shin2018medical,yang2018mri}. Yet, despite their visual realism, GAN-based methods often failed to preserve fine anatomical details and spatial consistency. More recently, diffusion models~\cite{ho2020denoising,song2020denoising,dhariwal2021diffusion,kazerouni2023diffusion} have emerged as a more stable alternative, producing high-fidelity and diverse images through likelihood-based training. Conditional diffusion models~\cite{xie2022measurement,lyu2022conversion,gungor2023adaptive,guo2025maisi} further improve structural control by conditioning on segmentation masks or metadata. However, most existing approaches operate in latent space or apply fixed intensity clipping, which compromises HU precision and weakens organ boundaries—limitations particularly critical for CT synthesis.

\begin{figure}[t]
\centering
    \includegraphics[width=\linewidth]{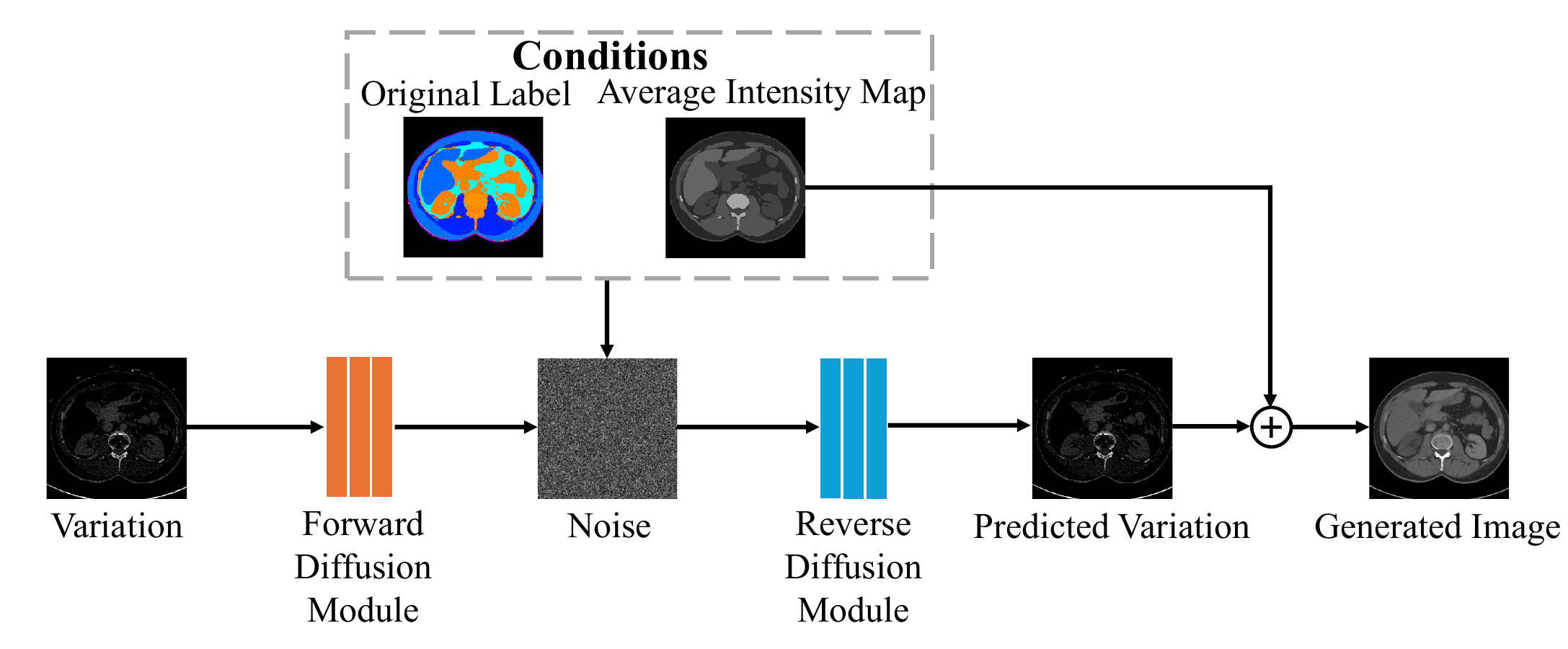}
    \caption{Variation Modeling via Diffusion for CT Image Synthesis. The variation is forward diffused to Gaussian noise. A reverse diffusion model, guided by the segmentation label and average intensity map, predicts the variation signal, which is added back to reconstruct the final CT image.}
    \label{fig:stage2}
\end{figure}

To address these issues, we propose the \emph{Prior-Integrated Variation Modeling} (PIVM) framework, as illustrated in Fig.~\ref{fig:stage2}. PIVM predicts voxel-level intensity deviations relative to organ-wise priors derived from segmentation labels. Operating entirely in image space, it preserves the full HU range and maintains anatomical alignment while capturing subtle texture variations. For volumetric synthesis, we extend this formulation to a sequential slice-wise generation strategy that ensures smooth 3D continuity across slices.

Our main contributions are as follows. (1) We introduce a prior-integrated diffusion formulation for anatomically faithful CT synthesis. (2) We design a lightweight slice-wise strategy for efficient 3D volume generation. (3) We demonstrate that the synthetic data generated by PIVM improves downstream segmentation accuracy on the TotalSegmentator dataset~\cite{wasserthal2023totalsegmentator}.

\begin{figure}[t]
\centering
    \includegraphics[width=0.7\linewidth]{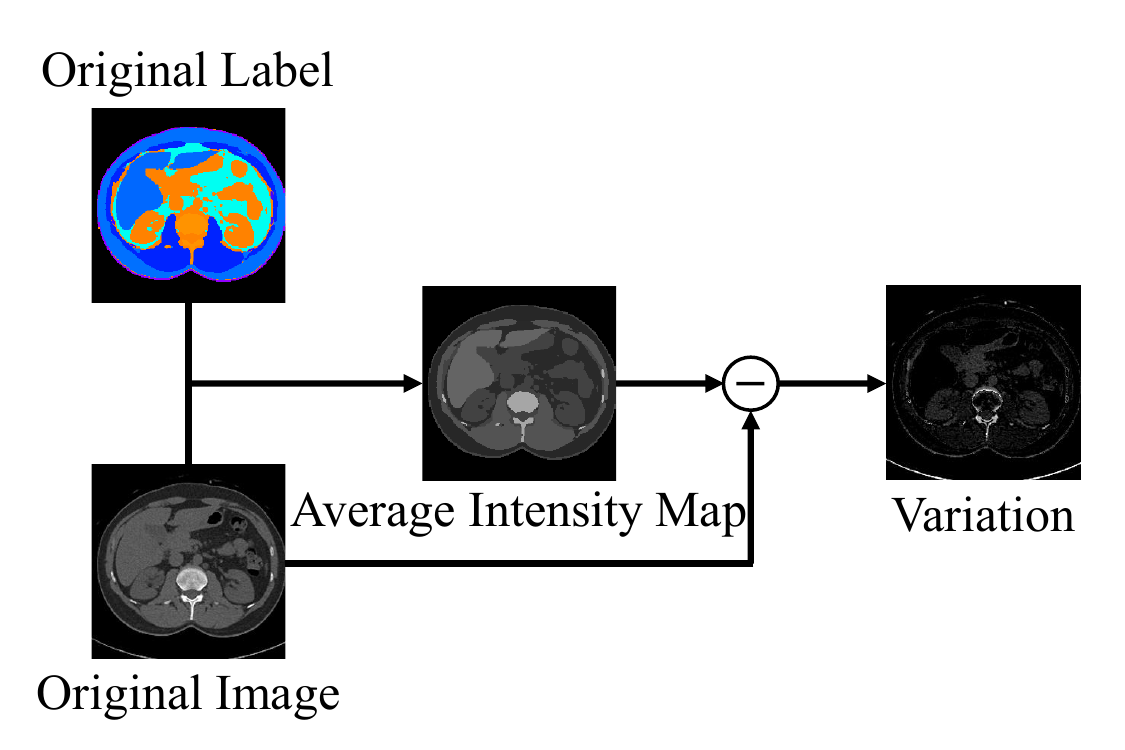}
    \caption{Construction of Average Intensity Map and Variation. An average intensity map is computed from the segmentation label by replacing each organ region with its dataset-level mean value. The pixel-wise difference between the original CT image and the average map defines the variation used in subsequent diffusion modeling.}
    \label{fig:stage1}
\end{figure}

\section{Method}
\label{sec:method}

We next describe the components of the PIVM framework introduced in Fig.~\ref{fig:stage2}. Our goal is to synthesize anatomically consistent and realistic CT images through the proposed PIVM framework. Given a CT slice \(x^{(i)}\), its segmentation map \(s^{(i)}\), and an average-intensity prior \(I^{(i)}\) encoding organ-level mean HU values, we define a residual variation
\[
v^{(i)} = x^{(i)} - I^{(i)}.
\]
An illustration of this construction is shown in Fig.~\ref{fig:stage1}, where the residual variation highlights organ-level deviations from the average intensity prior. Instead of generating absolute intensities, PIVM performs diffusion on this variation, focusing learning capacity on fine structural deviations that distinguish individual anatomy from its organ-level prior.

\subsection{Prior-Integrated Variation Modeling}

Denoising Diffusion Probabilistic Models (DDPMs)~\cite{ho2020denoising} synthesize data by progressively denoising Gaussian noise through a learned reverse process. In PIVM, diffusion is applied to the variation map \(v^{(i)}\) rather than the full image. During training, Gaussian noise is incrementally added over \(T\) steps, and the denoising network predicts the clean variation conditioned on both the segmentation label and intensity prior:
\[
\epsilon_\theta(v_t^{(i)}, t, s^{(i)}, I^{(i)}).
\]
After denoising, the final reconstruction is obtained as
\[
\hat{x}_0^{(i)} = I^{(i)} + \hat{v}_0^{(i)}.
\]
This formulation teaches the model to recover localized intensity deviations from anatomical averages, improving boundary precision and HU fidelity while remaining computationally efficient. A U-Net backbone serves as the denoising network, capturing both global context and fine-grained details via skip connections.

\subsection{Sequential Slice-Wise Volume Generation}

To generate full 3D volumes, PIVM is extended to a slice-wise synthesis strategy that maintains coherence across slices without the cost of full 3D diffusion. The first slice \(\hat{x}^{(1)}\) is generated from noise conditioned on its label \(s^{(1)}\) and prior \(I^{(1)}\). Each subsequent slice \(\hat{x}^{(t)}\) is produced using the previous slice as an additional conditioning input:
\[
\hat{x}^{(t)} = f_\theta(\hat{x}^{(t-1)}, s^{(t)}, I^{(t)}, \epsilon_t),
\]
where \(\epsilon_t \sim \mathcal{N}(0, I)\) represents Gaussian noise scaled to local intensity statistics. Conditioning on the prior slice encourages smooth transitions and consistent boundaries throughout the 3D volume, implicitly enforcing long-range anatomical continuity. This design achieves a strong balance between realism and efficiency, providing high structural fidelity at a fraction of the memory cost of volumetric diffusion models.

\section{Experiments and Results}

\subsection{Dataset and Preparation}
We trained and evaluated PIVM using the \textit{TotalSegmentator} dataset~\cite{wasserthal2023totalsegmentator}, which comprises 1,228 clinical CT volumes annotated for 117 anatomical structures.  
For this study, 865 abdominal scans were selected, each containing between 100 and 600 axial slices. To balance anatomical coverage and computational load, we extracted 12,853 representative 2D slices and standardized all images to \(256\times256\) pixels through resampling and zero-padding. Each slice was paired with segmentation masks for 34 major organs, including the liver, spleen, kidneys, and pancreas.

All intensities were converted to HU using DICOM metadata and clipped to \([-200,300]\) solely for the construction of \emph{average-organ-intensity maps}. These maps encode coarse structural priors that guide the diffusion process. The original intensity range was later restored after residual prediction.

For each organ \(k\), we computed a global mean intensity \(\mu_k\) across the training data:
\[
\mu_k = \frac{\sum_i \sum_{(u,v)\in\Omega_k^{(i)}} x_{u,v}^{(i)}}{\sum_i |\Omega_k^{(i)}|},
\quad \text{where } \Omega_k^{(i)} = \{(u,v): s_{u,v}^{(i)} = k\}.
\]
Each slice-specific prior map \(I^{(i)}\) was then constructed as:
\[
I^{(i)}_{u,v} =
\begin{cases}
\mu_k, & \text{if } s^{(i)}_{u,v} = k,\\
0, & \text{otherwise.}
\end{cases}
\]
These priors preserve anatomical layout while providing an intensity baseline for the variation-prediction stage.

\subsection{Implementation Details}
The diffusion backbone adopted a U-Net architecture for the denoising network \(\epsilon_\theta(x_t, t)\), equipped with symmetric skip connections to capture both global context and fine anatomical detail. 
Training was performed on 12{,}853 abdominal CT slices using the Adam optimiser (\(\beta_1 = 0.9, \beta_2 = 0.99\)) with an initial learning rate of \(1\times10^{-4}\), halved every 20\,epochs.  
Each batch contained 32 slices, and convergence was typically reached by around the 200\textsuperscript{th} epoch.  
All experiments were executed on a single NVIDIA RTX~4090 GPU, and end-to-end training of the complete framework required approximately 36\,hours.

For qualitative comparison, we reproduced the MAISI framework~\cite{guo2025maisi} using its official MONAI implementation. 
All architectural and training configurations followed the released model without modification, except for dataset paths and output settings.  
MAISI was executed in segmentation-to-image mode with its default preprocessing, which applies fixed HU clipping to the range \([-1000, 1000]\).  

\begin{figure}[t]
\centering
\renewcommand{\arraystretch}{1}
\setlength{\tabcolsep}{1pt}
\begin{tabular}{cccc}
    \includegraphics[width=0.23\linewidth]{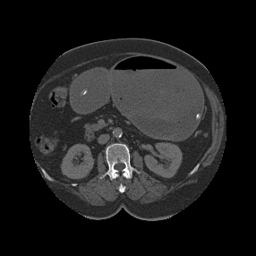} & 
    \includegraphics[width=0.23\linewidth]{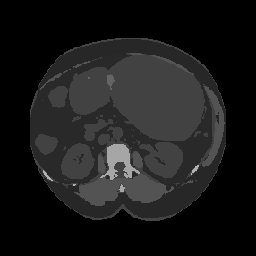} & 
    \includegraphics[width=0.23\linewidth]{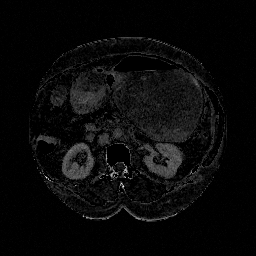} & 
    \includegraphics[width=0.23\linewidth]{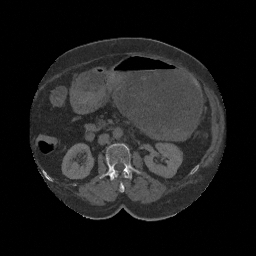} \\

    \includegraphics[width=0.23\linewidth]{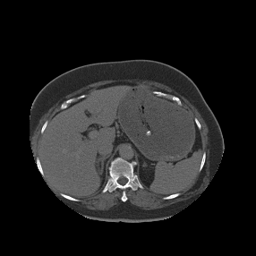} & 
    \includegraphics[width=0.23\linewidth]{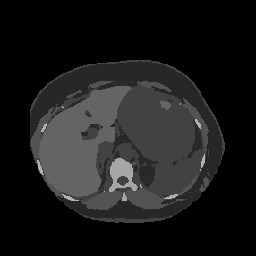} & 
    \includegraphics[width=0.23\linewidth]{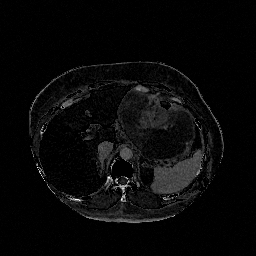} & 
    \includegraphics[width=0.23\linewidth]{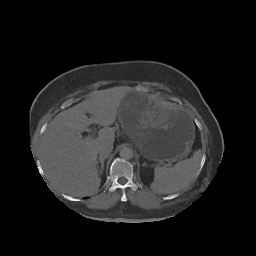} \\

    \includegraphics[width=0.23\linewidth]{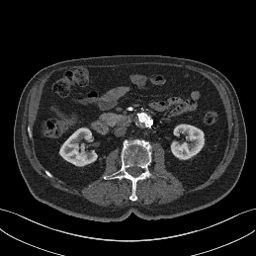} &
    \includegraphics[width=0.23\linewidth]{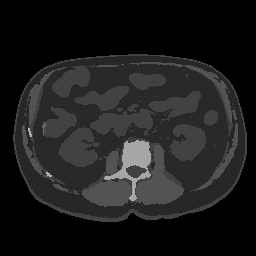} & 
    \includegraphics[width=0.23\linewidth]{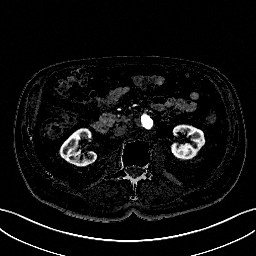} & 
    \includegraphics[width=0.23\linewidth]{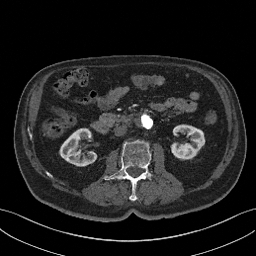} \\


\end{tabular}
\caption{Examples of PIVM’s generation process. Columns show the ground-truth CT image, the corresponding average organ intensity priors, the predicted intensity variations, and the reconstructed CT obtained by combining the prior with the variation.}
\label{fig:DDPM}
\end{figure}

\begin{figure}[!t]
\centering
\renewcommand{\arraystretch}{1}
\setlength{\tabcolsep}{1pt}
\begin{tabular}{ccccc}
    \includegraphics[width=0.19\linewidth]{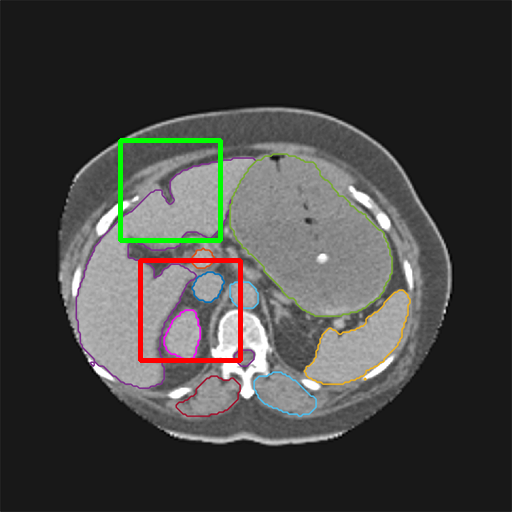} &
    \includegraphics[width=0.19\linewidth]{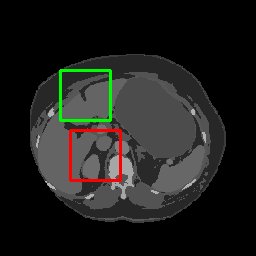} &
    \includegraphics[width=0.19\linewidth]{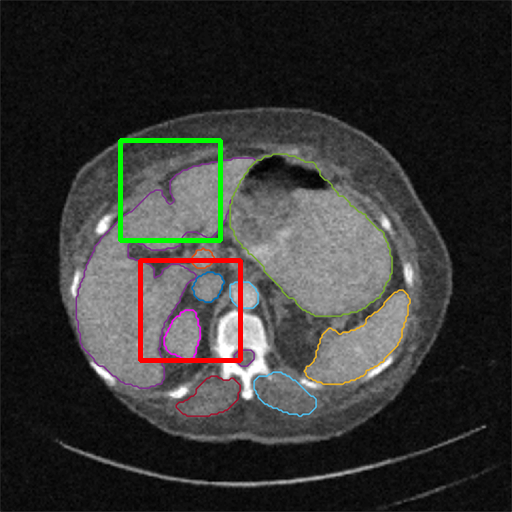} &
    \includegraphics[width=0.19\linewidth]{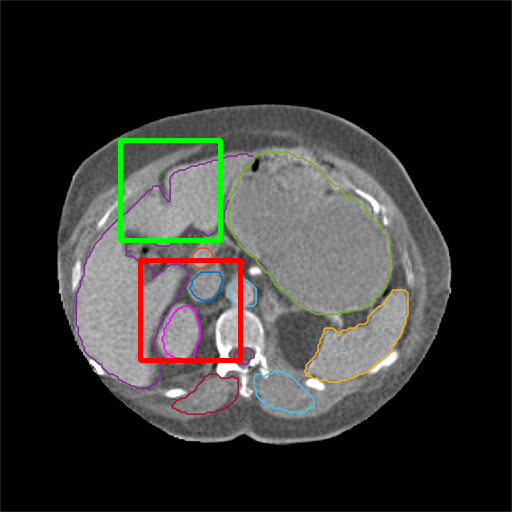} &
    \includegraphics[width=0.19\linewidth]{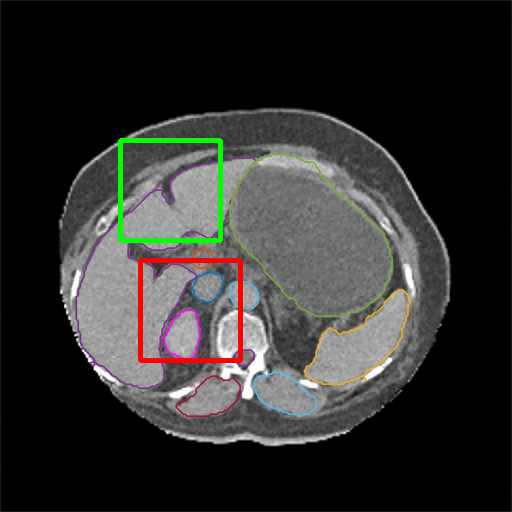} \\

    \includegraphics[width=0.19\linewidth]{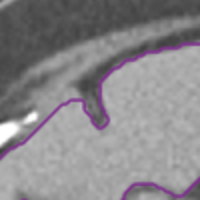} &
    \includegraphics[width=0.19\linewidth]{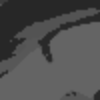} &
    \includegraphics[width=0.19\linewidth]{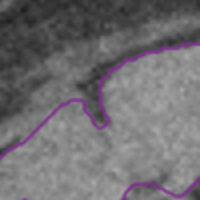} &
    \includegraphics[width=0.19\linewidth]{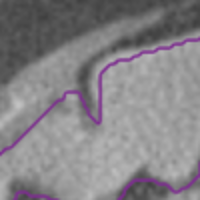} &
    \includegraphics[width=0.19\linewidth]{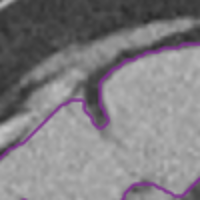} \\

    \includegraphics[width=0.19\linewidth]{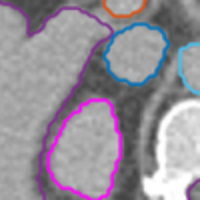} &
    \includegraphics[width=0.19\linewidth]{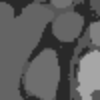} &
    \includegraphics[width=0.19\linewidth]{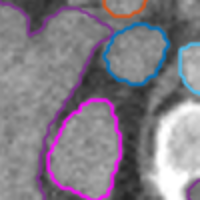} &
    \includegraphics[width=0.19\linewidth]{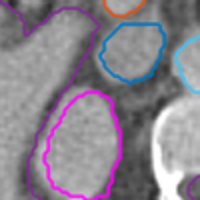} &
    \includegraphics[width=0.19\linewidth]{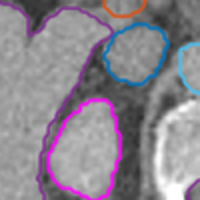} \\

    \makebox[0.19\linewidth][c]{(a) GT} &
    \makebox[0.19\linewidth][c]{(b) Label} &
    \makebox[0.19\linewidth][c]{(c) CDDPM} &
    \makebox[0.19\linewidth][c]{(d) MAISI} &
    \makebox[0.19\linewidth][c]{(e) PIVM}
\end{tabular}
\caption{Zoomed-in comparison of synthesis quality. Columns display: (a) ground-truth CT, (b) segmentation labels, (c) CDDPM outputs, (d) MAISI outputs, and (e) PIVM outputs.}
\label{fig:zoom}
\end{figure}

To assess image fidelity, we employed the Structural Similarity Index Measure (SSIM)~\cite{wang2004image} and the Fréchet Inception Distance (FID)~\cite{heusel2017gans}.  
For 3D volumes, visual inspection by experienced radiologists was used to evaluate spatial continuity and anatomical plausibility.

\subsection{Results}
\textbf{Quantitative Results.}
We use CDDPM~\cite{wang2025conditional} as our full-image diffusion baseline. 
The model operates directly in the image domain and predicts absolute HU values from noisy inputs using only the segmentation mask, without any organ-wise intensity priors.
Notably, CDDPM can be regarded as PIVM without the variation modeling step, since diffusion is applied directly to the full CT image. 
This setup provides a clear reference point for assessing the contribution of variation-based modeling in PIVM.
Representative qualitative results of PIVM are shown in Fig.~\ref{fig:DDPM}.
The synthesized CT images exhibit strong spatial alignment with the segmentation masks and preserve fine anatomical details.
As illustrated in Fig.~\ref{fig:zoom}, MAISI and CDDPM tend to generate over-smoothed textures due to latent-space sampling and HU clipping, while PIVM maintains subtle structural variations by operating directly in image space and preserving the full HU range.

\noindent
\textbf{Qualitative Results.} Quantitative results are summarized in Table~\ref{tab:results}.
PIVM achieves the best average reconstruction performance across five representative patients, with higher mean SSIM and lower mean FID than CDDPM and MAISI.

\begin{table*}[t]
\centering
\caption{Comparison of SSIM and FID among CDDPM, MAISI, and PIVM.}
\resizebox{0.75\textwidth}{!}{
\begin{tabular}{
>{\centering\arraybackslash}m{1.8cm}
c c c c c c}
\toprule
\multirow{2}{*}{Patient} & \multicolumn{3}{c}{SSIM $\uparrow$} & \multicolumn{3}{c}{FID $\downarrow$} \\ 
\cmidrule(lr){2-4} \cmidrule(lr){5-7}
& CDDPM & MAISI & PIVM (ours) & CDDPM & MAISI & PIVM (ours) \\
\midrule
1 & 0.231 & 0.742 & \textbf{0.846} & 117.66 & 133.70 & \textbf{63.12} \\
2 & 0.254 & 0.755 & \textbf{0.817} & 98.12 & \textbf{69.91} & 75.59 \\
3 & 0.364 & 0.638 & \textbf{0.716} & \textbf{70.09} & 117.91 & 72.71 \\
4 & 0.248 & 0.702 & \textbf{0.825} & 117.53 & 145.44 & \textbf{71.54} \\
5 & 0.280 & 0.705 & \textbf{0.799} & 91.61 & 121.49 & \textbf{54.12} \\
\midrule
\multicolumn{1}{c}{Avg. $\pm$ sd} &
0.275$\pm$0.053 &
0.708$\pm$0.046 &
\textbf{0.801$\pm$0.050} &
99.00$\pm$19.89 &
117.69$\pm$28.83 &
\textbf{67.42$\pm$8.76} \\
\bottomrule
\end{tabular}}
\label{tab:results}
\end{table*}

\begin{figure}[t]
\centering

\begin{minipage}[b]{0.45\linewidth}
    \centering
    \includegraphics[width=\linewidth]{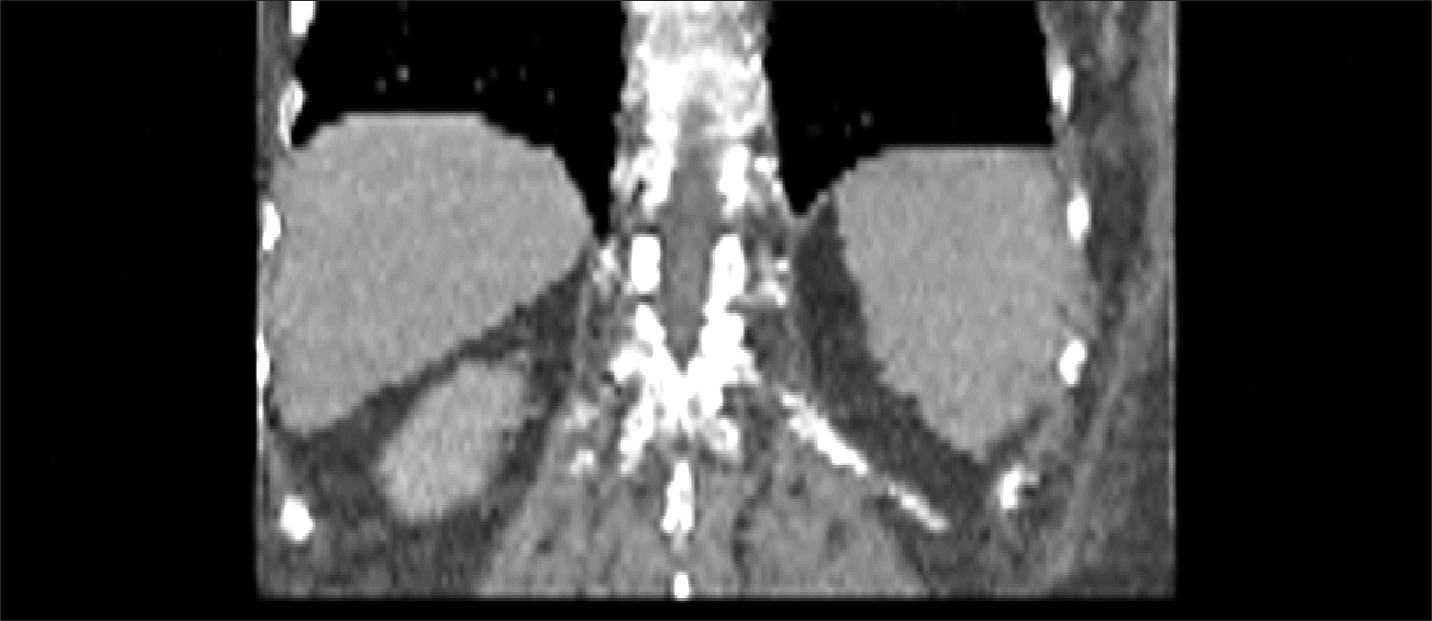}
    
    (a) Coronal view
\end{minipage}
\hfill
\begin{minipage}[b]{0.45\linewidth}
    \centering
    \includegraphics[width=\linewidth]{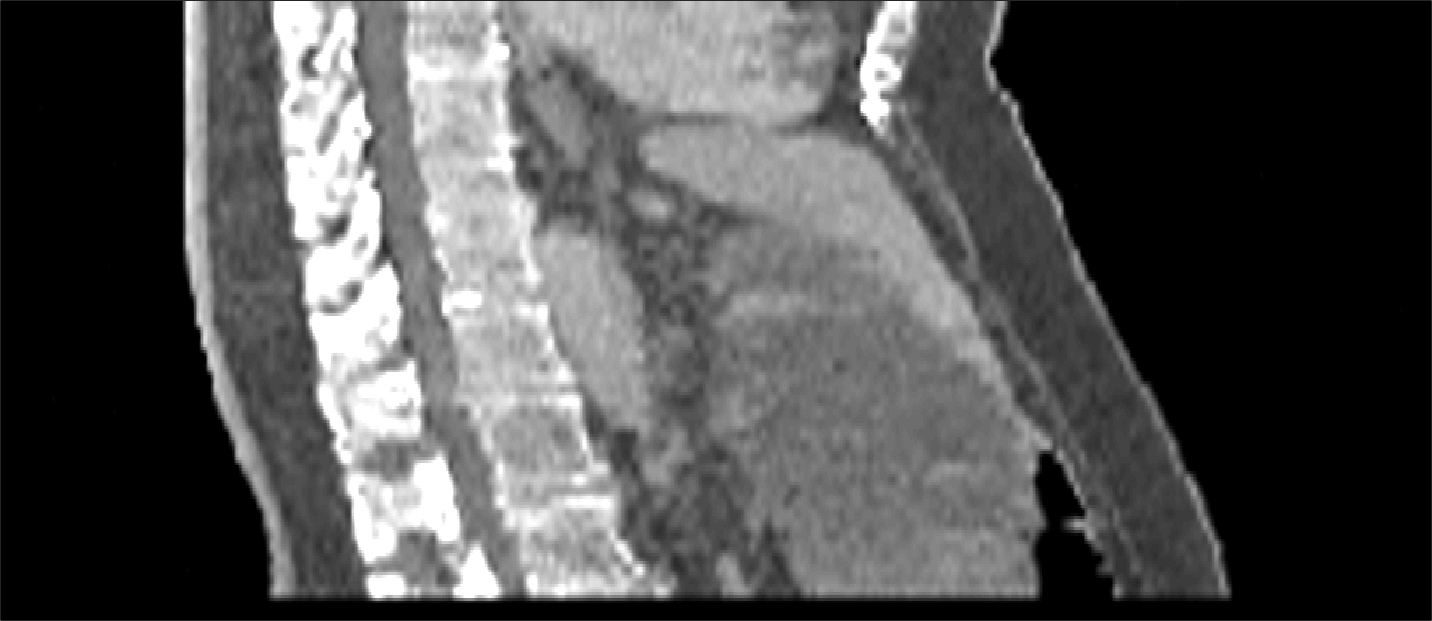}
    
    (b) Sagittal view
\end{minipage}

\caption{
Multi-view visualization of a generated 3D abdominal CT volume, illustrating anatomical fidelity and inter-slice continuity. Left: coronal view with overlaid multi-organ contours. Right: sagittal view of the same volume.
}
\label{fig:seq}
\end{figure}

\noindent
\textbf{3D Volume Visualization.} To evaluate volumetric coherence, we extended PIVM to sequentially synthesize 3D abdominal CT volumes slice by slice, as described in Section~\ref{sec:method}.  
Representative coronal and sagittal views of a generated volume are shown in Fig.~\ref{fig:seq}.  
The model preserves smooth inter-slice continuity and reconstructs major abdominal structures with realistic morphology, demonstrating that it effectively captures long-range spatial relationships.  
Expert inspection confirmed that the synthesized volumes exhibit anatomically plausible organ shapes and boundary transitions suitable for visualization, education, and training scenarios.

\subsection{Downstream Segmentation Task}

\begin{table}[htbp]
\small
\centering
\caption{Segmentation Performance Comparison for Models Trained on Datasets Generated by CDDPM, MAISI, and PIVM.}
\label{tab:seg}
\setlength{\tabcolsep}{6pt}
\begin{tabular}{lccc}
\toprule
Metric & CDDPM & MAISI & PIVM (ours) \\ \midrule
Hausdorff Distance (↓)    & 52.95  & 61.21 & \textbf{43.16} \\
Average Surface Dist. (↓) & 13.23 & 17.36 & \textbf{6.60}     \\
mIoU (↑)                  & 0.30 & 0.27 & \textbf{0.48}  \\
Dice Coefficient (↑)      & 0.38 & 0.34 & \textbf{0.53}     \\ 
\bottomrule
\end{tabular}
\end{table}

In Table~\ref{tab:seg}, we evaluate the use of PIVM-generated images in a downstream segmentation task. A U-Net model~\cite{ronneberger2015u} was trained on 5,000 synthetic slices produced by each method. Models trained with PIVM data consistently outperform those trained on CDDPM and MAISI across all metrics. This reflects improved boundary accuracy and region-level overlap.

We further examined the utility of PIVM-generated data in a downstream segmentation task with same segmentation configurations: 
(1) a real dataset of 5{,}000 annotated slices,  
(2) 5{,}000 synthetic slices generated by PIVM, and  
(3) a hybrid dataset combining both sources.  
As summarized in Table~\ref{tab:segmentation_performance}, the hybrid dataset yields the strongest overall performance. This demonstrates that PIVM-generated data provide effective augmentation for downstream segmentation.

\subsection{Ablation Study}
\begin{figure}[!t]
\small
\centering
\includegraphics[width=0.3\textwidth]{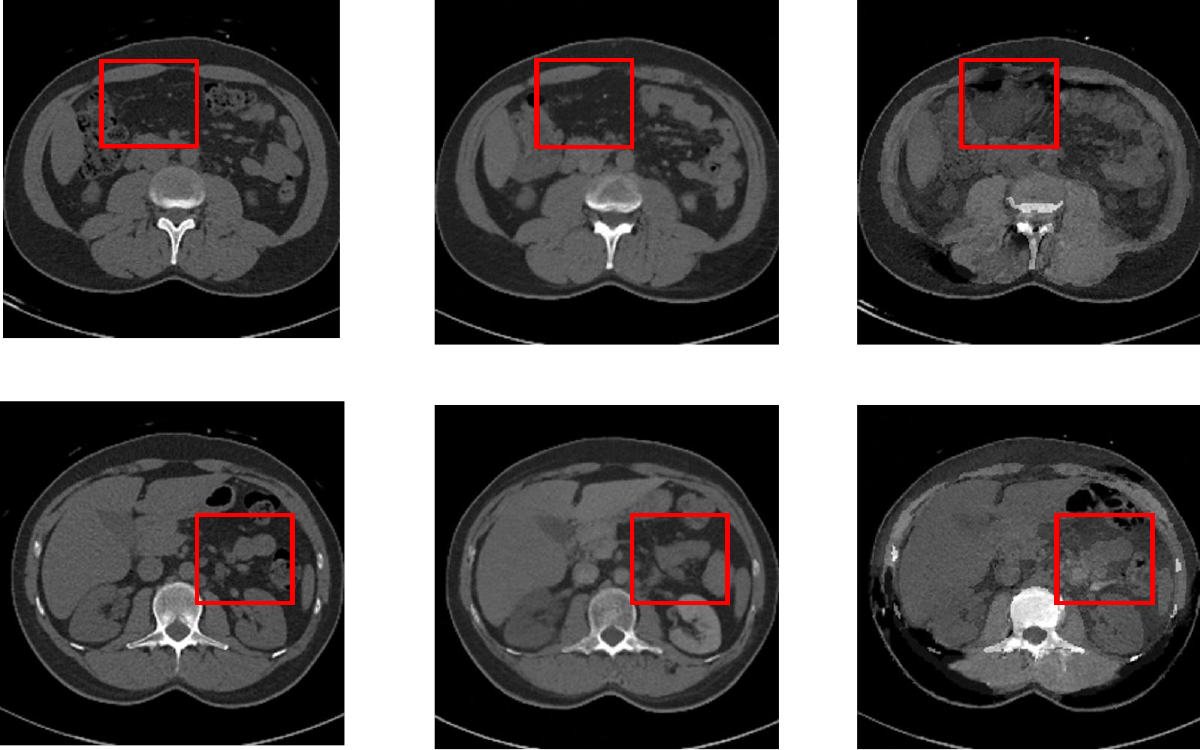}
\caption{Image generation results under different conditions. From left to right: ground truths (GT), images generated from \textit{34 labels}, and images generated from \textit{13 labels}.}
\label{fig:labels}
\end{figure}

To investigate the impact of segmentation label granularity on image synthesis quality, we varied the number of conditioning classes, as illustrated in Fig.~\ref{fig:labels}.
Using only 13 labels produced partially coherent structures but led to noticeable loss of fine details, particularly around skeletal regions.
In contrast, incorporating all 34 labels resulted in more realistic images with clearer anatomical boundaries and complete organ delineation.  

\begin{table}[htbp]
\small
\centering
\caption{Segmentation Performance for Models Trained on Original Dataset, Generated Dataset, and Combined Dataset.}
\label{tab:segmentation_performance}
\setlength{\tabcolsep}{3pt}
\resizebox{\linewidth}{!}{
\begin{tabular}{lccc}
\toprule
Metric & \shortstack{Original \\ Dataset} & \shortstack{Generated \\ Dataset} & \shortstack{Combined \\ Dataset} \\ \midrule
Hausdorff Distance (↓)    & 23.54  & 43.16 & \textbf{22.24} \\
mIoU (↑)                  & 0.78 & 0.48  & \textbf{0.81} \\
Dice Coefficient (↑)      & 0.87 & 0.53 & \textbf{0.88} \\ 
\bottomrule
\end{tabular}
}
\end{table}


\color{black}


\section{Conclusion}
\label{sec:discussion_conclusion}
This work presents PIVM, a diffusion-based approach to synthesising anatomically precise abdominal CT images. By modeling voxel-wise deviations from organ-specific intensity priors, the method retains the entire HU range and preserves the fine structural detail that is often blurred by conventional latent-space diffusion. Both visual assessment and quantitative evaluation show that PIVM produces smoother organ boundaries and more realistic tissue contrasts, supporting its potential for data augmentation and simulation in medical imaging. However, some residual artefacts, particularly in small vessels and diaphragmatic regions, reflect the limitations of current label resolution and sampling balance. Future improvements may come from finer-grained conditioning and fully volumetric generation. Overall, PIVM shows that incorporating explicit anatomical information into diffusion modeling can produce synthetic CT data that are both realistic and clinically interpretable.

\section{Compliance with Ethical Standards}

This study was conducted using fully anonymised, publicly available human CT data from open-access sources. According to the data usage licence, formal ethical approval was not required.

\section{ACKNOWLEDGMENTS}
This research was supported by the National Institute of Biomedical Imaging and Bioengineering (NIBIB) under grant R01-EB029431, with partial support from the Department of Radiation Oncology, Washington University School of Medicine in St. Louis.

{\small
\bibliographystyle{ieee_fullname}
\bibliography{egbib}

@article{ho2020denoising,
  title={Denoising diffusion probabilistic models},
  author={Ho, Jonathan and Jain, Ajay and Abbeel, Pieter},
  journal={Advances in neural information processing systems},
  volume={33},
  pages={6840--6851},
  year={2020}
}

@article{heusel2017gans,
  title={Gans trained by a two time-scale update rule converge to a local nash equilibrium},
  author={Heusel, Martin and Ramsauer, Hubert and Unterthiner, Thomas and Nessler, Bernhard and Hochreiter, Sepp},
  journal={Advances in neural information processing systems},
  volume={30},
  year={2017}
}

@article{song2020denoising,
  title={Denoising diffusion implicit models},
  author={Song, Jiaming and Meng, Chenlin and Ermon, Stefano},
  journal={arXiv preprint arXiv:2010.02502},
  year={2020}
}

@article{goodfellow2014generative,
  title={Generative adversarial nets},
  author={Goodfellow, Ian and Pouget-Abadie, Jean and Mirza, Mehdi and Xu, Bing and Warde-Farley, David and Ozair, Sherjil and Courville, Aaron and Bengio, Yoshua},
  journal={Advances in neural information processing systems},
  volume={27},
  year={2014}
}

@inproceedings{mao2017least,
  title={Least squares generative adversarial networks},
  author={Mao, Xudong and Li, Qing and Xie, Haoran and Lau, Raymond YK and Wang, Zhen and Paul Smolley, Stephen},
  booktitle={Proceedings of the IEEE international conference on computer vision},
  pages={2794--2802},
  year={2017}
}

@article{gulrajani2017improved,
  title={Improved training of wasserstein gans},
  author={Gulrajani, Ishaan and Ahmed, Faruk and Arjovsky, Martin and Dumoulin, Vincent and Courville, Aaron C},
  journal={Advances in neural information processing systems},
  volume={30},
  year={2017}
}

@inproceedings{shin2018medical,
  title={Medical image synthesis for data augmentation and anonymization using generative adversarial networks},
  author={Shin, Hoo-Chang and Tenenholtz, Neil A and Rogers, Jameson K and Schwarz, Christopher G and Senjem, Matthew L and Gunter, Jeffrey L and Andriole, Katherine P and Michalski, Mark},
  booktitle={Simulation and Synthesis in Medical Imaging: Third International Workshop, SASHIMI 2018, Held in Conjunction with MICCAI 2018, Granada, Spain, September 16, 2018, Proceedings 3},
  pages={1--11},
  year={2018},
  organization={Springer}
}

@misc{dhariwal2021diffusion,
      title={Diffusion Models Beat GANs on Image Synthesis}, 
      author={Prafulla Dhariwal and Alex Nichol},
      year={2021},
      eprint={2105.05233},
      archivePrefix={arXiv},
      primaryClass={cs.LG}
}

@article{lyu2022conversion,
  title={Conversion between CT and MRI images using diffusion and score-matching models},
  author={Lyu, Qing and Wang, Ge},
  journal={arXiv preprint arXiv:2209.12104},
  year={2022}
}

@inproceedings{xie2022measurement,
  title={Measurement-conditioned denoising diffusion probabilistic model for under-sampled medical image reconstruction},
  author={Xie, Yutong and Li, Quanzheng},
  booktitle={International Conference on Medical Image Computing and Computer-Assisted Intervention},
  pages={655--664},
  year={2022},
  organization={Springer}
}

@article{kazerouni2023diffusion,
  title={Diffusion models in medical imaging: A comprehensive survey},
  author={Kazerouni, Amirhossein and Aghdam, Ehsan Khodapanah and Heidari, Moein and Azad, Reza and Fayyaz, Mohsen and Hacihaliloglu, Ilker and Merhof, Dorit},
  journal={Medical Image Analysis},
  pages={102846},
  year={2023},
  publisher={Elsevier}
}

@article{gungor2023adaptive,
  title={Adaptive diffusion priors for accelerated MRI reconstruction},
  author={G{\"u}ng{\"o}r, Alper and Dar, Salman UH and {\"O}zt{\"u}rk, {\c{S}}aban and Korkmaz, Yilmaz and Bedel, Hasan A and Elmas, Gokberk and Ozbey, Muzaffer and {\c{C}}ukur, Tolga},
  journal={Medical Image Analysis},
  volume={88},
  pages={102872},
  year={2023},
  publisher={Elsevier}
}

@article{zhu2017unpaired,
  title={Unpaired image-to-image translation using cycle-consistent adversarial networks},
  author={Zhu, Jun-Yan and Park, Taesung and Isola, Phillip and Efros, Alexei A},
  journal={Proceedings of the IEEE international conference on computer vision},
  pages={2223--2232},
  year={2017}
}

@article{yang2018mri,
  title={MRI to CT synthesis using deep convolutional networks},
  author={Yang, Xi and others},
  journal={Proceedings of the International Conference on Medical Image Computing and Computer-Assisted Intervention (MICCAI)},
  pages={169--177},
  year={2018}
}

@article{wasserthal2023totalsegmentator,
  title={TotalSegmentator: robust segmentation of 104 anatomic structures in CT images},
  author={Wasserthal, Jakob and Breit, Hanns-Christian and Meyer, Manfred T and Pradella, Maurice and Hinck, Daniel and Sauter, Alexander W and Heye, Tobias and Boll, Daniel T and Cyriac, Joshy and Yang, Shan and others},
  journal={Radiology: Artificial Intelligence},
  volume={5},
  number={5},
  year={2023},
  publisher={Radiological Society of North America}
}

@article{de2022ct,
  title={CT study protocol optimization in acute non-traumatic abdominal settings},
  author={De Muzio, F and Cutolo, C and Granata, V and Fusco, R and Ravo, L and Maggialetti, N and Brunese, MC and Grassi, R and Grassi, F and Bruno, F and others},
  journal={Eur. Rev. Med. Pharmacol. Sci},
  volume={26},
  pages={860--878},
  year={2022}
}

@article{rhee2021role,
  title={The role of imaging in current treatment strategies for pancreatic adenocarcinoma},
  author={Rhee, Hyungjin and Park, Mi-Suk},
  journal={Korean Journal of Radiology},
  volume={22},
  number={1},
  pages={23},
  year={2021},
  publisher={Korean Society of Radiology}
}

@inproceedings{deheyab2022overview,
  title={An overview of challenges in medical image processing},
  author={Deheyab, A Omar Adil and Alwan, Mohammed Hasan and Rezzaqe, Islam khalid Abdul and Mahmood, Omar Abdulkareem and Hammadi, Yousif I and Kareem, Ali Noori and Ibrahim, Maha},
  booktitle={Proceedings of the 6th International Conference on Future Networks \& Distributed Systems},
  pages={511--516},
  year={2022}
}

@article{litjens2017survey,
  title={A survey on deep learning in medical image analysis},
  author={Litjens, Geert and Kooi, Thijs and Bejnordi, Babak Ehteshami and Setio, Arnaud Arindra Adiyoso and Ciompi, Francesco and Ghafoorian, Mohsen and Van Der Laak, Jeroen Awm and Van Ginneken, Bram and S{\'a}nchez, Clara I},
  journal={Medical image analysis},
  volume={42},
  pages={60--88},
  year={2017},
  publisher={Elsevier}
}

@inproceedings{ronneberger2015u,
  title={U-net: Convolutional networks for biomedical image segmentation},
  author={Ronneberger, Olaf and Fischer, Philipp and Brox, Thomas},
  booktitle={Medical image computing and computer-assisted intervention--MICCAI 2015: 18th international conference, Munich, Germany, October 5-9, 2015, proceedings, part III 18},
  pages={234--241},
  year={2015},
  organization={Springer}
}

@article{ma2021abdomenct,
  title={Abdomenct-1k: Is abdominal organ segmentation a solved problem?},
  author={Ma, Jun and Zhang, Yao and Gu, Song and Zhu, Cheng and Ge, Cheng and Zhang, Yichi and An, Xingle and Wang, Congcong and Wang, Qiyuan and Liu, Xin and others},
  journal={IEEE Transactions on Pattern Analysis and Machine Intelligence},
  volume={44},
  number={10},
  pages={6695--6714},
  year={2021},
  publisher={IEEE}
}

@inproceedings{guo2025maisi,
  title={Maisi: Medical ai for synthetic imaging},
  author={Guo, Pengfei and Zhao, Can and Yang, Dong and Xu, Ziyue and Nath, Vishwesh and Tang, Yucheng and Simon, Benjamin and Belue, Mason and Harmon, Stephanie and Turkbey, Baris and others},
  booktitle={2025 IEEE/CVF Winter Conference on Applications of Computer Vision (WACV)},
  pages={4430--4441},
  year={2025},
  organization={IEEE}
}

@inproceedings{wang2025conditional,
  title={Conditional Diffusion Model for Abdominal CT Image Synthesis},
  author={Wang, Xu and He, Dinglun and Zhang, Baoming and Hao, Yao and Yang, Deshan and Duan, Ye},
  booktitle={2025 IEEE 22nd International Symposium on Biomedical Imaging (ISBI)},
  pages={1--5},
  year={2025},
  organization={IEEE}
}

@article{wang2004image,
  title={Image quality assessment: from error visibility to structural similarity},
  author={Wang, Zhou and Bovik, Alan C and Sheikh, Hamid R and Simoncelli, Eero P},
  journal={IEEE transactions on image processing},
  volume={13},
  number={4},
  pages={600--612},
  year={2004},
  publisher={IEEE}
}
}

\end{document}